\documentclass[lettersize,journal]{IEEEtran}
\usepackage{array}
\usepackage[caption=false,font=normalsize,labelfont=sf,textfont=sf]{subfig}
\usepackage{textcomp}
\usepackage{stfloats}
\usepackage{url}
\usepackage{verbatim}
\usepackage{graphicx}
\usepackage{hyperref}
\usepackage{cite}

\usepackage{colortbl}
\usepackage{pifont}
\usepackage[noend]{algpseudocode}  
\usepackage{booktabs}
\usepackage{multirow}
\usepackage{amsmath,amssymb,amsfonts}
\usepackage[ruled,vlined,linesnumbered,boxed,commentsnumbered]{algorithm2e}

\def\etal{\emph{et al.}}

\def\BibTeX{{\rm B\kern-.05em{\sc i\kern-.025em b}\kern-.08em
    T\kern-.1667em\lower.7ex\hbox{E}\kern-.125emX}}
\usepackage{balance}
\begin{document}
\title{FedPalm: A General Federated Learning Framework for Closed- and Open-Set Palmprint Verification}
\author{Ziyuan~Yang, Yingyu Chen, Chengrui~Gao, Andrew~Beng~Jin~Teoh,~\IEEEmembership{Senior Member,~IEEE,}\\
Bob~Zhang,~\IEEEmembership{Senior Member,~IEEE,} Yi Zhang,~\IEEEmembership{Senior Member,~IEEE}
        

\thanks{This work has been submitted to the IEEE for possible publication. Copyright may be transferred without notice, after which this version may no longer be accessible. (\textit{Corresponding Authors: Andrew Beng Jin Teoh and Yi Zhang})}
\thanks{Ziyuan Yang is with the College of Computer Science, Sichuan University, Chengdu 610065, China, and also with Centre for Frontier AI Research~(CFAR), Agency for Science, Technology and Research~(A*STAR), Singapore (e-mail: cziyuanyang@gmail.com)}
\thanks{Yingyu Chen and Chengrui Gao are with the College of Computer Science, Sichuan University, Chengdu 610065, China (e-mail: cyy262511@gmail.com and cr@stu.scu.edu.cn)}
\thanks{Andrew Beng Jin Teoh is with the School of Electrical and Electronic Engineering, College of Engineering, Yonsei University, Seoul, Republic of Korea (e-mail: bjteoh@yonsei.ac.kr)}
\thanks{Bob Zhang is with the Pattern Analysis and Machine Intelligence Group, Department of Computer and Information Science, University of Macau, Taipa, Macau, China (e-mail: bobzhang@um.edu.mo)}
\thanks{Yi Zhang is with the School of Cyber Science and Engineering, Sichuan University, Chengdu 610065, China (e-mail: yzhang@scu.edu.cn)}}


\markboth{Journal of \LaTeX\ Class Files,~Vol.~18, No.~9, September~2020}%
{How to Use the IEEEtran \LaTeX \ Templates}

\maketitle

\begin{abstract}

Current deep learning (DL)-based palmprint verification models rely on centralized training with large datasets, which raises significant privacy concerns due to biometric data's sensitive and immutable nature. Federated learning~(FL), a privacy-preserving distributed learning paradigm, offers a compelling alternative by enabling collaborative model training without the need for data sharing. However, FL-based palmprint verification faces critical challenges, including data heterogeneity from diverse identities and the absence of standardized evaluation benchmarks. This paper addresses these gaps by establishing a comprehensive benchmark for FL-based palmprint verification, which explicitly defines and evaluates two practical scenarios: closed-set and open-set verification. We propose FedPalm, a unified FL framework that balances local adaptability with global generalization. Each client trains a personalized textural expert tailored to local data and collaboratively contributes to a shared global textural expert for extracting generalized features. To further enhance verification performance, we introduce a Textural Expert Interaction Module that dynamically routes textural features among experts to generate refined side textural features. Learnable parameters are employed to model relationships between original and side features, fostering cross-texture-expert interaction and improving feature discrimination. Extensive experiments validate the effectiveness of FedPalm, demonstrating robust performance across both scenarios and providing a promising foundation for advancing FL-based palmprint verification research. The related code will be publicly available upon acceptance at \url{https://github.com/Zi-YuanYang/FedPalm}.

\end{abstract}


\section{Introduction}
\IEEEPARstart{B}{iometrics} plays a crucial role in verification and identification systems. Biometric traits are generally classified into physiological and behavioral characteristics. Due to the stability and reliability of physiological traits, biometrics such as face, iris, and fingerprint recognition have demonstrated promising performance and achieved widespread adoption in daily life~\cite{jain2004introduction}. Among these technologies, palmprint recognition has garnered significant attention due to its rich texture patterns, ease of acquisition, and high security~\cite{fei2018feature}.

Over the years, palmprint-based biometric systems have advanced significantly, with traditional methods focusing on textural feature extraction for verification. Zhang \textit{et al.}~\cite{zhang2003online} introduced Gabor filters to capture texture features, inspiring further studies on texture extractors~\cite{jia2008palmprint}, coding rules~\cite{yang20212TCC}, and competitive mechanisms~\cite{xu2018drcc}. However, these methods often rely heavily on domain expertise to design suitable extractors, which limits their adaptability.

Inspired by the success of deep learning~(DL), recent efforts have combined traditional methods with DL techniques. Early attempts, such as Dian \textit{et al.}~\cite{dian2016contactless} adapting AlexNet, faced challenges due to palmprint fine-grained issues such as intra-class variations and inter-class similarities. Recent works have improved performance, including learnable Gabor filters~\cite{liang2021compnet} and competition-based networks~\cite{yang2023ccnet}. However, the data-intensive nature of DL raises privacy concerns, as large-scale palmprint data collection risks exposing sensitive, immutable biometric information, posing significant security challenges~\cite{yan2024toward}.

Federated learning (FL) has emerged as a transformative approach for distributed learning. It enables collaborative model training across multiple clients without exchanging data. FL significantly enhances privacy preservation by keeping sensitive biometric data localized to each client. This addresses a critical challenge in palmprint recognition, where data leakage could lead to irreversible security breaches. 

Beyond privacy, FL offers additional motivations for its application in palmprint recognition. First, FL optimizes resources by leveraging distributed edge devices' computational power and data diversity. Second, it fosters robustness and adaptability, as models trained on diverse, decentralized datasets can better generalize to unseen scenarios. Third, FL promotes compliance with regulations like GDPR~\cite{voigt2017eu}, emphasizing data sovereignty and localized processing. For instance, in a cross-institutional biometric authentication system, FL allows hospitals or banks to train a palmprint verification model while preserving client confidentiality collaboratively. Similarly, FL benefits wearable devices, where palmprint data collected locally on smartwatches or handheld scanners can contribute to a global model without exposure, ensuring privacy and scalability.

Recognizing these advantages, researchers have begun integrating palmprint verification methods into the FL framework. For example, Shao \textit{et al.}~\cite{shao2024palmfedmetric} tackled the non-IID (non-independent and identically distributed) data issue in palmprint systems by utilizing a globally shared dataset across clients. Similarly, Yang \textit{et al.}~\cite{yang2024physics} introduced an innovative FL approach that ensures spectral consistency, improving model performance without compromising privacy.


Despite advances in the field, a well-defined experimental benchmark for FL in palmprint verification remains absent. This paper bridges this gap by establishing clear tasks and standardized evaluation protocols for federated palmprint verification. Specifically, we identify two primary scenarios: closed-set and open-set settings. The closed-set scenario assumes that all identities in the test set are also in the training set. In contrast, the open-set scenario introduces unseen identities during testing, posing greater challenges for generalization and robustness. Both scenarios should be evaluated together, representing distinct and practical application settings.

Furthermore, we require that clients operate independently, without sharing identities or relying on globally shared real data. Allowing overlapping identities among clients is impractical, and using shared global data risks compromising individual privacy, ultimately undermining the core objective of privacy preservation in FL. Moreover, inherent differences in textures among individuals exacerbate the non-IID issue, posing significant challenges for maintaining consistency across clients.

DL models often tend to overfit local datasets, a challenge particularly pronounced in biometric verification. Overfitting leads to significant variations in optimization directions across clients. Traditional FL methods~\cite{mcmahan2017communication} aggregate these diverse optimization directions to update a global model. However, this approach often compromises performance on local data, as the aggregated model struggles to adapt to client-specific nuances. Conversely, personalized FL methods aim to fit local datasets better, but their performance often deteriorates when confronted with unseen data~\cite{sabah2024model}. 

Interestingly, these contrasting approaches mirror the dynamics of closed-set and open-set scenarios in biometrics. Generic FL methods align with closed-set settings, where models perform well on predefined categories but struggle to generalize. In contrast, personalized FL methods resonate with open-set scenarios, focusing on adaptability to local data at the expense of broader generalization. Balancing these approaches is crucial to addressing the unique challenges of biometric verification in FL systems.


Building on this insight, we introduce \textbf{FedPalm}, a unified FL framework designed to handle both closed-set and open-set scenarios in palmprint recognition. FedPalm achieves this by training both personalized local and shared global models. Each client maintains a personalized model for closed-set recognition while contributing to a globally shared model optimized for open-set scenarios. The FedPalm framework divides the recognition network into textural expert layers and embedding layers. This architecture enables each client to use two models: one personalized for the closed-set setting and another linked to the global model for open-set recognition. By integrating these specialized models, FedPalm ensures robust performance across diverse recognition scenarios, balancing local adaptability with global generalization.


The personalized closed-set models are tailored to local data, delivering superior performance on client-specific tasks while mitigating the non-IID issue inherent to FL. In addition to these personalized models, each client trains an aggregated model to form the globally shared model. This global model ensures robustness and generalizability, especially for open-set tasks, addressing the diverse requirements of biometric verification.


To further exploit the strengths of both model types and extract more discriminative features, we introduce the Textural Expert Interaction Module (TEIM). Specifically, for each model, including personalized closed-set and global shared open-set models, the TEIM routes different textural features extracted from the other textural experts to generate side textural features. Then, we employ learnable parameters to model the relationship between the original and neighboring textural features, enabling the model's embedding layer to extract a comprehensive and robust textural representation. This approach mitigates overfitting by leveraging diverse textural features to enhance generalization. Additionally, routing the most relevant textural features improves robustness and minimizes the potential negative impact caused by the significant data.



The main contributions of this paper are as follows:
\begin{itemize}
    \item We establish a comprehensive benchmark for federated learning in palmprint verification. We explicitly define closed-set and open-set settings to address diverse application needs and provide standardized evaluation protocols for real-world scenarios.
    \item We propose FedPalm, a unified FL framework that performs well in closed and open-set settings.
    \item We propose TEIM, designed to enhance robustness and discriminative power by dynamically blending original textural features with those from neighboring sources.
\end{itemize}

\section{Related Works}

\subsection{Palmprint Verification}
Palmprint verification has undergone significant advancements across different technological milestones. In the early stages, researchers proposed numerous traditional methods that heavily relied on prior knowledge. These methods are broadly categorized into statistic-based and coding-based approaches~\cite{fei2020feature}.

Coding-based methods, particularly, demonstrated remarkable performance by effectively capturing textural features. For example, Guo~\etal~\cite{guo2009palmprint} introduced the Binary Orientation Co-occurrence Vector~(BOCV), which used six Gabor filters to extract and encode magnitude features. Later, Yang~\etal~\cite{yang20212TCC} extended BOCV by incorporating multi-order texture features to improve performance.
In addition, Kong~\etal~\cite{kong2003palmprint} proposed a competition mechanism to extract order features that demonstrated robustness to lighting variations. Inspired by the success of order features in palmprint verification, many follow-up studies refined the performance and applicability of these methods~\cite{fei2016double,yang2020extreme,fei2016half}.

Despite their success, traditional approaches rely heavily on handcrafted features and domain expertise, limiting their adaptability and scalability. With the advent of DL, palmprint recognition has entered a new era. For instance, Zhong~\etal~\cite{zhong2018palmprint} introduced a palmprint recognition method using a siamese network. Similarly, Chai~\etal~\cite{chai2019boosting} leveraged DL networks to enhance recognition performance by incorporating gender information. Genovese~\etal~\cite{genovese2019palmnet} proposed PalmNet, which combined features extracted from multiple feature extractors to improve recognition accuracy. Hashing networks have also gained attention in this field~\cite{jia2021deep,wu2022multi,dong2022co}.

Recently, researchers have sought to integrate traditional methods with DL technologies. For example, CompNet~\cite{liang2021compnet} and CCNet~\cite{yang2023ccnet} extend traditional coding-based methods. Inspired by these works, numerous learnable Gabor filters~(LGF)-based methods have been proposed for palmprint recognition~\cite{yang2023co3net,gao2024cross,yu2023multi}.
Although these approaches demonstrate promising performance, they largely overlook privacy concerns. Most of these methods assume that clients can collect large-scale data to train powerful models, which may not be practical in privacy-preserving scenarios.


\subsection{Federated Learning}
FL was first introduced as a distributed learning paradigm in \cite{mcmahan2017communication}. These methods are divided into generic and personalized FL methods. Generic FL methods aim to train a global shared model that covers all client-side data. For example, Li~\etal~\cite{li2020federated} introduced the proximal loss to constrain the optimization directions of local models to avoid model shift issues. Wang~\etal~\cite{wang2020tackling} proposed a novel optimizer to address this issue. Additionally, some researchers introduced a contrastive item between local and global models to address non-iid issues, such as MOON~\cite{li2021model} and FedKD~\cite{wu2022communication}.

However, these methods often fail to fit local data well, leading to poor performance on clients with data distributions significantly different from others~\cite{zhang2023fedala}. Researchers have recently proposed personalized FL approaches to enhance local performance~\cite{yang2023dynamic}. For example, FedPer~\cite{arivazhagan2019federated} creates specific classifiers, sharing some global layers.

Li~\etal~\cite{li2021fedbn} proposed FedBN, which alleviated heterogeneity concerns by personalizing batch normalization~(BN) layers. Some works utilized class prototype learning in personalized FL to achieve satisfactory performance~\cite{huang2023rethinking,cheng2023protohar,dai2023tackling}. Additionally, some works introduced side information to improve performance~\cite{yang2023hypernetwork,chen2024hyperfednet,shin2024effective}.

Several works have also introduced FL into palmprint recognition. For example, Shao~\etal~\cite{shao2023privacy} proposed FedML, which alleviates the non-IID issue by leveraging a globally shared public dataset and employing metric learning. Similarly, in~\cite{shao2020towards}, the authors utilized a global shared dataset to address data heterogeneity. Yang~\etal~\cite{yang2024physics} assumed that each client shares the same set of individuals and proposed a novel FL paradigm to achieve spectrum consistency.

Despite growing interest in this field, a comprehensive and practical benchmark for federated palmprint recognition remains a significant gap. This hinders progress and limits its potential to drive future research.

\section{Benchmark Setting in FL-based Palmprint Recognition}

In this paper, we present a benchmark for FL-based palmprint recognition that covers both \textbf{closed-set} and \textbf{open-set} scenarios, enabling a comprehensive evaluation of performance across diverse application settings. To this end, we explicitly define the two primary settings for FL-based palmprint recognition:

\subsubsection{Closed-Set Palmprint Recognition}
closed-set palmprint recognition assumes that enrolled (gallery) and testing identities (query) are predefined and included in the training set. This scenario aims for highly accurate recognition performance within a fixed group of known individuals. Since all identities are available during training, the task is to match palmprints to specific identities within this predetermined group accurately. The closed-set setting is particularly useful in controlled environments, such as access control systems or enterprise-level authentication, where the identity pool is fixed and predefined.

\subsubsection{Open-Set Palmprint Recognition}
In contrast, open-set palmprint recognition introduces a more complex and practical challenge by assuming that the enrolled (gallery) and testing (query) identities are entirely separate from the training set. This scenario requires the recognition system to handle palmprints from unseen individuals, making the task much more challenging. Open-set recognition aligns with real-world scenarios like public security systems and surveillance, where the system must identify new individuals not part of the training data.


\subsubsection{Federated Learning Constraints}
Due to the distributed nature of FL in palmprint verification, the benchmark follows  strict privacy-preserving constraints as follows:

\begin{itemize}
\item \textbf{No Global Shared Real Data:} Clients are prohibited from sharing any globally accessible real dataset, ensuring that privacy is preserved without centralized data aggregation.

\item \textbf{Non-IID Data Distribution:} Each client's data distribution may differ significantly, reflecting real-world scenarios where clients have diverse and heterogeneous palmprint datasets.

\item \textbf{Identity Isolation:} Clients cannot share overlapping identities, ensuring a realistic setting where each client operates independently with unique identity pools.
\end{itemize}

This benchmark provides a foundation for evaluating FL-based palmprint recognition systems, addressing challenges related to privacy, data heterogeneity, and generalization across unseen identities. It serves as a practical framework for developing and testing algorithms under realistic constraints, paving the way for future advancements in evaluating privacy-preserving biometric recognition methods.

\begin{figure*}
    \centering
    \includegraphics[width=\linewidth]{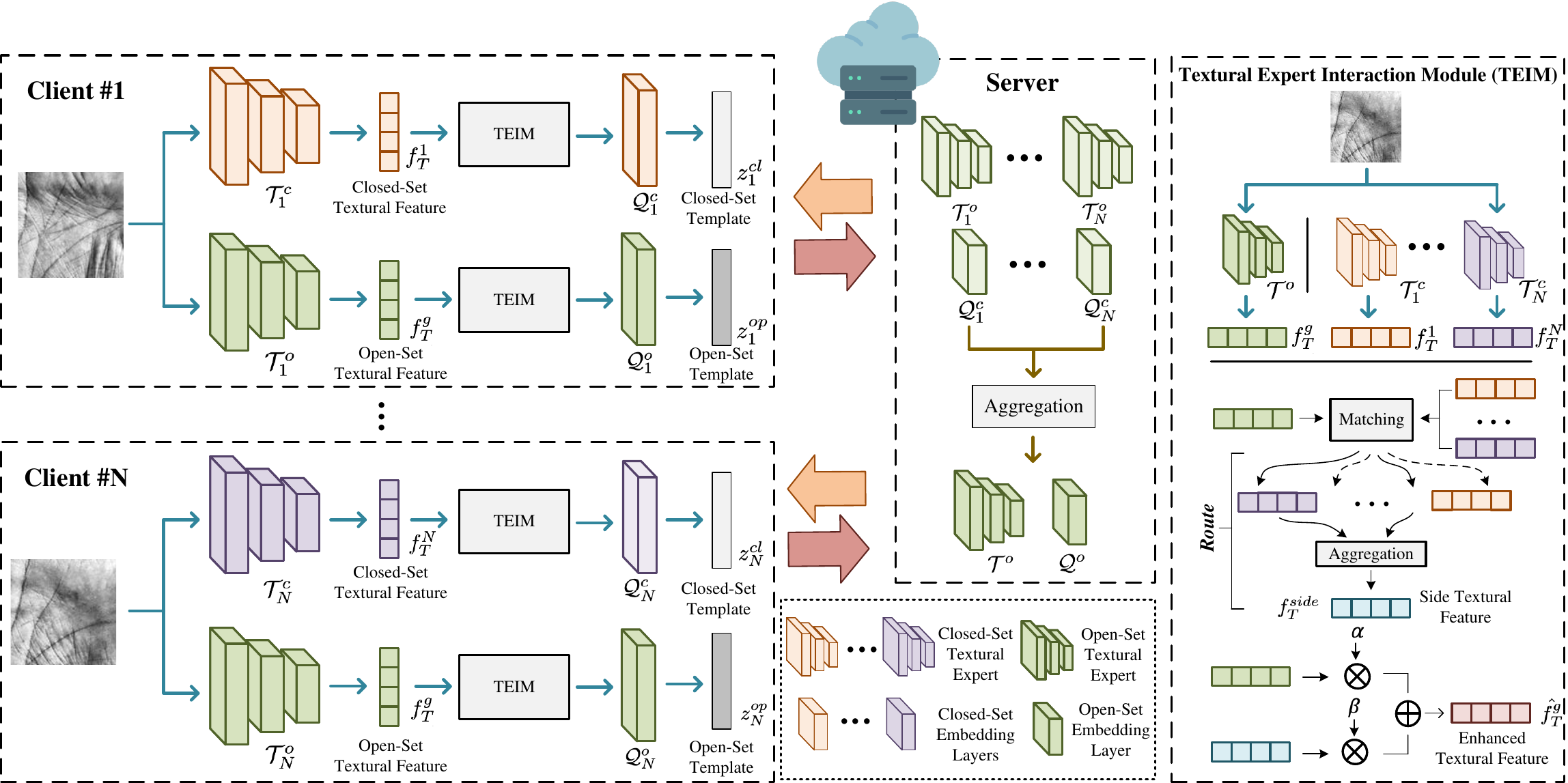}
      \caption{Overview of the proposed FedPalm. In the proposed TEIM, $N$ local textural experts and one shared global textural expert collaboratively extract features from the input image. TEIM routes features from other experts to generate enhanced features for each expert, which are then used for subsequent embedding.}
  \label{fig:Over} 
\end{figure*}

\section{Methodology}

\subsection{Overview}
The pipeline of the proposed FedPalm framework is depicted in Fig.~\ref{fig:Over}. In FedPalm, each client trains two recognition models: one optimized for client-specific tasks in a closed-set setting and another designed for generalizability in an open-set setting.

To address the non-IID issue and promote texture diversity, we propose personalizing the closed-set models for each client, including the textural experts and embedding layers. Additionally, all local open-set models are aggregated, allowing the \textit{global shared open-set model} to extract generalized textural features that enhance overall performance.

To improve feature discriminability, we introduce TEIM, a novel feature interaction component in FedPalm. TEIM integrates local and global textural features to facilitate cross-expert interactions and refine feature representations. By dynamically routing relevant features and capturing their relationships, TEIM enhances robustness and discriminative power while filtering irrelevant information, delivering superior performance across diverse scenarios.


\subsection{Learning Paradigm}
In the FedPalm framework, each client trains two distinct palmprint verification models one tailored for the closed-set task and the other for the open-set task. In the following discussion, these models will be referred to as the \textit{closed-set} and \textit{open-set models} for clarity.

Previous studies have highlighted the critical role of textural features in palmprint verification. Building on this insight, we propose structuring the palmprint verification models into two components: a textural expert and an embedding layer.

In the closed-set setting, FedPalm personalizes the textural expert in the closed-set model to align with the unique characteristics of local data. This approach addresses challenges such as non-IID data distribution and identity isolation. Furthermore, personalization enriches textural diversity, a critical factor in enhancing discriminability and robustness in subsequent processing stages.

Specifically, we personalize the closed-set models \textit{without}, including them in the aggregation process. This design ensures that each closed-set model remains dedicated to local data and unaffected by information from other clients. The models capture diverse textural features across clients, enabling each expert to focus exclusively on local data. This approach establishes a robust foundation for the subsequent steps in our framework.

However, while personalization in closed-set models enhances local adaptation, it limits their ability to capture global textural features. To address this limitation, we propose that each client trains an open-set model alongside the closed-set model, enabling the extraction of local and global features.

The open-set model plays a critical role in capturing global textural knowledge. It aims to achieve the best average performance across all clients, ensuring generalization to unseen data. The global model learns shared patterns and features by aggregating knowledge from all clients, preventing overfitting any single client's data. This aggregation process promotes robust performance in open-set scenarios by balancing generalization and specificity.

In the FedPalm framework, all open-set model components, such as the textural expert and embedding layers, are aggregated after each communication round. This aggregation follows the classical method in federated learning~\cite{mcmahan2017communication}, ensuring consistency among clients and scalability across the system.

While this dual-model approach addresses closed-set and open-set settings, the two models operate independently, leaving room for further optimization. We introduce TEIM, which connects the two models to bridge this gap. By fostering collaboration and knowledge sharing, TEIM enhances the overall learning process and ensures superior performance across diverse scenarios.

\subsection{Textural Expert Interaction Module}

The FedPalm framework introduces a collaborative architecture with $N+1$ models: $N$ personalized closed-set models and a single globally shared open-set model. As a result, FedPalm leverages the expertise of $N+1$ specialized textual experts to achieve its objectives.

For each palmprint image, we first use the $N+1$ textural experts to extract textural features, defined as follows:

\begin{equation}
F_T=\{f_T^1,...,f_T^N,f_T^g\},
\end{equation}
where $F_T$ denotes the set of all textural features. $f_T^g$ represents the global textural features from the open-set model, and $f_T^i (i=1,2,...,N)$ represents the local textural features from the closed-set models. 

To clarify, we define the relevant variables to streamline the subsequent discussion. The closed-set and open-set models of the $i$-th client are represented as $\mathcal{M}_i^{c}$ and $\mathcal{M}^{o}$, respectively. Each palmprint verification model comprises a textural expert, $\mathcal{T}$, and embedding layers, $\mathcal{Q}$. Specifically, $\mathcal{T}_i^c$ and $\mathcal{T}^o$ denote the textural experts of $\mathcal{M}_i^{c}$ and $\mathcal{M}^{o}$, respectively. $\mathcal{Q}_i^c$ and $\mathcal{Q}^o$ denote the embedding layers of $\mathcal{M}_i^{c}$ and $\mathcal{M}^{o}$. Additionally, $z_i^{cl}$ and $z_i^{op}$ denote the closed-set and open-set feature templates for the $i$-th client, respectively. Next, $f_T^g$ and $f_T^i$ are the outputs of $\mathcal{T}^o$ and $\mathcal{T}_i^o$, respectively.

The module routes the most relevant textural features from the remaining $N$ experts to generate the \textit{side textural features} for each textural expert. Specifically:
\begin{itemize}
    \item For the global open-set model, the most relevant textural features are selected from $F_T \setminus \{f_T^g\}$.
    \item For the $i$-th textural expert, the most relevant textural features are selected from $F_T \setminus \{f_T^i\}$.
\end{itemize}

Then, we employ a unified routing paradigm to select the relevant textural features for simultaneously closed- and open-set models. For simplicity, the following discussion will focus on operations performed on the global shared open-set model. The same processes apply to the closed-set models as well.

To leverage the diverse knowledge acquired from various clients, the proposed TEIM dynamically routes the most relevant textual features. This begins by assessing the relationships among different textual features, which can be expressed as follows:
\begin{equation}
    d_i = 1 - dis(f_T^g, f_T^i), i=1,...,N
\end{equation}
where $d_i$ represent the similarity between $f_T^g$ and $f_T^i$. 

To enhance verification performance, we propose a method that routes the top three most similar textural features identified by the other experts. These features are averaged to generate a side textural feature, denoted as $f_T^{side}$. By leveraging features from multiple clients, the model captures complementary information, significantly improving its representational capacity compared to using local features alone. The process is expressed as follows:  

\begin{equation}  
    IN = Route(d_0, \ldots, d_N),  
\end{equation}  
where $Route$ is the algorithm used to rank the indices of the most relevant features in descending order based on their relevance, denoted as $IN$. Using $IN$, the side textural feature is computed as follows:  

\begin{equation}  
f_T^{side} = \frac{1}{K} \sum_{i=1}^{K} f_T^{IN(i)},  
\label{eq:side}
\end{equation}  
where $K$ represents the number of top-ranked features selected, and $f_T^{IN(i)}$ denotes the textural feature corresponding to the $i$-th ranked index in $IN$. We recommend setting $K=3$ as the default configuration for optimal performance based on empirical evaluation.  

This approach effectively integrates insights from multiple sources, enriching the model's feature representation and boosting overall verification accuracy.  
This process of discarding less relevant features also mitigates the challenges posed by the non-IID nature of federated data by filtering out significant disparities between textural features. As a result, only the most relevant and consistent features from diverse clients are aggregated, improving the model's generalizability in a federated setting. 

To further improve performance, we introduce learnable parameters $\{\alpha, \beta\}$, which dynamically aggregate $f_T^g$ and $f_T^{side}$ to yield an enhanced textural feature $\hat{f_T^{g}}$. This process can be formally expressed as follows:
\begin{equation}
    \hat{f_T^{g}} = \alpha f_T^g + \beta f_T^{side}.
\end{equation} 

These parameters adaptively adjust the contribution of each feature based on its importance, which enables the model to balance local and cross-client information effectively. By leveraging these learnable parameters, the model better captures the intricate relationships between the original and side textural features, ensuring that the aggregated representation remains discriminative. The enhanced feature $\hat{f_T^{g}}$ is then passed through the embedding layer to produce the final feature representation, which serves as the template. 

While this example is described in the context of the global shared open-set model, the same applies to closed-set experts, yielding $\hat{f_T^{i}} = \alpha f_T^i + \beta f_T^{side}$. This unified approach ensures consistency across all models, both global and local. To avoid redundancy, we will not elaborate separately on the process for the local closed-set models, as it follows the same pipeline.

\subsection{Implementation}
\subsubsection{Training Details} 

To streamline the following discussion, we define the relevant notations and variables. For the $i$-th client, the closed-set and open-set models are denoted as $\mathcal{M}_i^{c}$ and $\mathcal{M}_i^{o}$, respectively. Each palmprint verification model consists of a textural expert, $\mathcal{T}$, and embedding layers, $\mathcal{Q}$. 


As depicted in Fig.~\ref{fig:Over}, the training process in FedPalm personalizes the closed-set models to adapt to client-specific data while enabling the aggregation of other models. The aggregation process of the open-set textural expert $\mathcal{T}^o$ at the server can be formulated as:
\begin{equation}
        \theta_\mathcal{T}^o =  \frac{\sum_{i=1}^{N} n_{i} {\theta_\mathcal{T}}_i^o}{\sum_{i=1}^{N} n_{i}},
        \label{eq:agg_tex}
\end{equation}
where $n_i$ denotes the number of samples in the $i$-th client and ${\theta_\mathcal{T}}_i^o$ denotes the parameters of $\mathcal{M}_i^{o}$. 

Similarly, the aggregation of the open-set embedding layers $\mathcal{Q^o}$ at the server can be formulated as:

\begin{equation}
\theta_\mathcal{Q}^o=\frac{\sum_{i=1}^{N} n_{i} {\theta_\mathcal{Q}}_i^o}{\sum_{i=1}^{N} n_{i}},
\label{eq:emb}
\end{equation}
where ${\theta_\mathcal{Q}}_i^o$ denotes the parameters of $\mathcal{Q}_i^{o}$.

In this paper, we utilize the hybrid loss proposed in~\cite{yang2023ccnet} to optimize the local models, which consist of the cross-entropy loss and the contrastive loss. The cross-entropy loss is formulated as:
\begin{equation}
    \mathcal{L}_{ce} = -\frac{1}{n}\sum_{i=1}^n\sum_{c=1}^{M} y_{i,c} \log \left(p_{i,c}\right),
\end{equation}
where $n$ and $M$ denote the numbers of samples and classes, respectively. $y_{i,c}$ and $p_{i,c}$ represent the label and the predicted probability of the $i$-th sample.

The contrastive loss is formulated as:
\begin{equation}
    \mathcal{L}_{con} = -\sum_{i \in I} \frac{1}{|P(i)|} \sum_{p \in P(i)} \log \frac{\exp \left(z_{i} \cdot z_{p} / \tau\right)}{\sum_{a \in A(i)} \exp \left(z_{i} \cdot z_{a} / \tau\right)},
\end{equation}
where $I\equiv\{1 \ldots 2 n\}$ is the batch of contrastive sample pairs, $A(i) \equiv I \backslash\{i\}$ and $i$ is the index of the positive sample. $P(i)\equiv {p \in A(i): y_i = y_p}$ is the index set of the positive samples in the batch, distinct from $i$, and $y_i$ is the label of the $i$-th sample in the batch. $|P(i)|$ stands for the number of samples in $P(i)$. $z_{i}$ and $z_{p}$ are the anchor and the positive template features derived from the enhanced textural feature. $\tau$ is the temperature factor. 

Finally, the hybrid loss function is formulated as follows:
\begin{equation}
    \mathcal{L}= w_{ce} \mathcal{L}_{ce} + w_{con} \mathcal{L}_{con},
    \label{eq:loss}
\end{equation}
where $w_{ce}$ and $w_{con}$ are the weights of the cross-entropy and contrastive losses, respectively. We follow previous works to set them to 0.8 and 0.2, respectively~\cite{yang2023ccnet}.

During training, the TEIM is deliberately kept inactive during the first third of all communication rounds. This intentional delay allows the textural experts to converge independently, ensuring stability before introducing potentially inconsistent features from other clients. Premature integration of these external textural features at this stage could disrupt the training process, leading to instability and slower progress.

In parallel, updates to the closed-set textural experts are also restricted to the initial one-third phase of the communication rounds. Once these experts stabilize, they remain fixed for the remainder of the training. This approach minimizes fluctuations in local textural features, providing a stable foundation for the TEIM. This strategy facilitates the seamless convergence of subsequent embedding layers by ensuring consistent inputs, thus promoting a more robust and efficient training process.

For a comprehensive understanding of our proposed method, the main steps of the FedPalm framework are outlined in Algorithm~\ref{alg:PSFed-Palm}. TEIM($\cdot$) denotes the proposed Textural Expert Interaction Module. Additionally, $p_i^{cl},z_i^{cl},f_T^i$ represent the prediction, feature template, and textural feature of the closed-set model $\mathcal{M}_i^c$. Similarly, $p_i^{op},z_i^{op},{f_T}_i^g$ denote the corresponding components of the open-set model $\mathcal{M}_i^o$.

\begin{algorithm}[t]
  \caption{Main steps of the FedPalm.}  
  \label{alg:PSFed-Palm}
   \textbf{Definition:} $R$ represents the number of communication rounds, $N$ denotes the number of clients, $D^i$ denotes the dataset in the $i$-th client, and $E$ denotes the number of local epochs. $x$ and $y$ denote the palmprint image and corresponding label, respectively. 
   
   
   \textbf{Function Main:} \Comment{Server Executes}
   
    Initialize ${\theta_\mathcal{T}}^{c,0},{\theta_\mathcal{T}}^{o,0},{\theta_\mathcal{Q}}^{c,0},{\theta_\mathcal{Q}}^{c,0}$ and send them to clients.
    
    \For{round $r=1,2,...,R$}{
    
        \For{client $i=1,2,...,N$ \rm{\textbf{in parallel}}}{${\theta_\mathcal{T}}_i^{o,r},{\theta_\mathcal{Q}}_i^{o,r} \gets$ \textbf{Local Training}($r,i, \theta_\mathcal{T}^{o,r},\theta_\mathcal{Q}^{o,r})$
       \\
       $\theta_\mathcal{T}^{o,r+1} \gets$ Aggregation(${\theta_\mathcal{T}}_1^{o,r},...,{\theta_\mathcal{T}}_i^{o,r}$) \par \Comment{Based on Eq.~\eqref{eq:agg_tex}} \\
        $\theta_\mathcal{Q}^{o,r+1} \gets$ Aggregation(${\theta_\mathcal{Q}}_1^{o,r},...,{\theta_\mathcal{Q}}_i^{o,r}$) \par \Comment{Based on Eq.~\eqref{eq:emb}} \\
        Send ${\theta_\mathcal{Q}}^{o,r+1}$ and ${\theta_\mathcal{T}}^{o,r+1}$ to clients.
    }
    }
    \textbf{Local Training}($r,i,{\theta_\mathcal{Q}}^{o,r}, {\theta_\mathcal{T}}^{o,r}$): \Comment{Client Executes}\\
        \For{$e=1,2,...,E$}{
        
        \For{$(x, y)$ in $D^i$}{
            $p_i^{cl}, z_i^{cl}, f_T^i \gets \mathcal{M}_i^{c}(x,y;{\theta_\mathcal{Q}}_i^{c,r},{\theta_\mathcal{T}}^{c,r})$\\
            $p_i^{op}, z_i^{op}, {f_T}_i^{g}\gets \mathcal{M}_i^{o}(x,y;{\theta_\mathcal{Q}}_i^{o,r},{\theta_\mathcal{T}}_i^{o,r})$

            \textbf{if}($r<R / 3$):

            \quad  Calculate the loss and optimize ${\theta_\mathcal{Q}}_i^{c,r},{\theta_\mathcal{T}}_i^{c,r},{\theta_\mathcal{Q}}^{o,r},{\theta_\mathcal{T}}^{o,r}$
            \Comment{Based on Eq.~\eqref{eq:loss}}

            \textbf{else:}
                
            \quad $\hat{f_T^{i}} \gets$ TEIM$(f_T^i,x,F_T \setminus \{f_T^i\})$

            \quad $p_{cl}, z_{cl} \gets \mathcal{Q}_i^c(\hat{f_T^{i}};{\theta_\mathcal{Q}}_i^{c,r})$

            \quad $\hat{f_T^{g}} \gets$ TEIM$(f_T^g,x,F_T \setminus \{f_T^g\})$

            \quad $p_{op}, z_{op} \gets \mathcal{Q}_i^c(\hat{f_T^{g}};{\theta_\mathcal{Q}}^{o,r})$

            \quad  Calculate the loss and optimize ${\theta_\mathcal{Q}}_i^{c,r},{\theta_\mathcal{Q}}^{o,r}$, and ${\theta_\mathcal{T}}^{o,r}$
            \Comment{Based on Eq.~\eqref{eq:loss}}
        }
        }
        \textbf{return} ${\theta_\mathcal{Q}}^{o,r}$, and ${\theta_\mathcal{T}}^{o,r}$

\end{algorithm}

\subsubsection{Deployment}

After training, each client uploads its textural expert $\mathcal{T}$ to the server, which subsequently redistributes these experts to all $N$ clients. Consequently, each client receives $N-1$ textural expert from other clients, along with its personalized closed-set $\mathcal{M}_i^c$ and the globally shared open-set models $\mathcal{M}^o$.

In the closed-set scenario, an input image is processed through the $N-1$ textural experts, $\mathcal{M}_i^c$ and $\mathcal{M}^o$. To clarify, we define the textural feature extracted by the client’s personalized closed-set model as the anchor textural feature $f_a^c$. TEIM then dynamically routes the most relevant features of $f_a^c$ from $F_T$, generating a side textural feature, $f_a^{cside}$. The $f_a^c$ is then enhanced by integrating $f_a^{cside}$ via $\hat{f^{c}} = \alpha f_a^c + \beta f_a^{cside}$. The $\hat{f^c}$ is then passed through the embedding layer of the client’s personalized closed-set model to produce the feature template $z^c=\mathcal{Q}^c_i(\hat{f^c})$.

A similar process is followed for the open-set scenario. The primary distinction lies in the source of the anchor textural feature, which, in this case, is extracted from the globally shared open-set model, $\mathcal{M}^o$. The enhanced feature $\hat{f}_T^g$ is then processed through the embedding layer of the open-set model, $\mathcal{Q}^o$. This unified framework ensures a consistent and seamless approach to both closed-set and open-set scenarios, enabling effective adaptation to diverse application needs.

The open-set model is tailored for dynamic and practical scenarios where the system must handle new or unseen individuals. This is particularly valuable in applications such as public security systems or large-scale deployments, where enrolling all potential users during training is unfeasible.


Together with the closed-set model, these complementary capabilities enable FedPalm to adapt seamlessly to diverse and evolving use cases.

\renewcommand{\arraystretch}{1.1}
\begin{table*}[!t]
\centering
\caption{The Closed-Set EER Performance in Tongji.}
\resizebox{.85\textwidth}{!}{
\begin{tabular}{lccccccccc}
\hline
\multicolumn{1}{c}{} & Client \#1 & Client \#2 & Client \#3 & Client \#4 & Client \#5 & Client \#6 & Client \#7 & \multicolumn{1}{c|}{Client \#8} & Average \\ \hline
w/o FL               & 0.94\%     & 1.33\%     & 0.32\%     & 0.37\%     & 1.14\%     & 0.46\%     & 0.73\%     & \multicolumn{1}{c|}{0.95\%}     & 0.78\%  \\ \hline
\multicolumn{10}{c}{Generic FL Methods}                                                  \\
\hline 
FedAvg               & 1.45\%     & 2.05\%     & 0.89\%     & 0.97\%     & 1.62\%     & 0.78\%     & 1.38\%     & \multicolumn{1}{c|}{1.03\%}     & 1.27\%  \\
FedProx              & 1.74\%     & 2.28\%     & 1.24\%     & 1.30\%     & 1.95\%     & 0.93\%     & 1.59\%     & \multicolumn{1}{c|}{1.22\%}     & 1.53\%  \\
FedNova              & 7.39\%     & 8.97\%     & 7.92\%     & 7.79\%     & 7.41\%     & 7.86\%     & 7.96\%     & \multicolumn{1}{c|}{6.46\%}     & 7.72\%  \\
MOON                 & 1.05\%     & 1.26\%     & 1.10\%     & 0.97\%     & 0.72\%     & 1.46\%     & 0.94\%     & \multicolumn{1}{c|}{1.19\%}     & 1.09\%  \\
FedKD                & 0.63\%     & 1.71\%     & 2.21\%     & 0.96\%     & 1.73\%     & 0.76\%     & 0.70\%     & \multicolumn{1}{c|}{1.73\%}     & 1.30\%  \\
\hline
\multicolumn{10}{c}{Personazlied FL Methods} 
 \\   \hline
FedPer               & 0.37\%     & 0.92\%     & 0.32\%     & 0.34\%     & 1.59\%     & 0.43\%     & 1.70\%     & \multicolumn{1}{c|}{0.54\%}     & 0.78\%  \\
FedBN                & 0.88\%     & 1.66\%     & 0.74\%     & 2.00\%     & 1.84\%     & 2.11\%     & 1.08\%     & \multicolumn{1}{c|}{0.57\%}     & 1.36\%  \\
HyperFed             & 0.84\%     & 0.32\%     & 1.47\%     & 0.50\%     & 0.57\%     & 0.46\%     & 1.25\%     & \multicolumn{1}{c|}{0.35\%}     & 0.72\%  \\ \hline
FedPalm              & 0.53\%     & 0.96\%     & 1.08\%     & 0.18\%     & 0.27\%     & 0.72\%     & 0.22\%     & \multicolumn{1}{c|}{0.89\%}     & \textbf{0.61\% } \\ \hline
\end{tabular}
}
\label{tab:cl_tongji}
\end{table*}

\begin{table*}[!t]
\centering
\caption{The Open-Set EER Performance in Tongji Dataset (Personalized FL Methods).}
\resizebox{.85\textwidth}{!}{
\begin{tabular}{lccccccccc}
\hline
\multicolumn{1}{c}{Method} & Client \#1 & Client \#2 & Client \#3 & Client \#4 & Client \#5 & Client \#6 & Client \#7 & \multicolumn{1}{c|}{Client \#8} & Average  \\ \hline
\multicolumn{10}{c}{Non-FL Method}  \\ \hline
w/o FL               & 10.72\%    & 10.91\%    & 10.57\%    & 10.97\%    & 10.56\%    & 11.62\%    & 10.71\%    & \multicolumn{1}{c|}{11.03\%}    & 10.89\% \\ \hline
\multicolumn{10}{c}{Personazlied FL Methods} \\ \hline
FedPer               & 10.66\%    & 10.40\%    & 11.61\%    & 10.75\%    & 11.13\%    & 10.66\%    & 10.41\%    & \multicolumn{1}{c|}{11.07\%}    & 10.84\% \\
FedBN                & 11.99\%    & 12.88\%    & 12.74\%    & 12.25\%    & 12.16\%    & 12.13\%    & 12.79\%    & \multicolumn{1}{c|}{12.32\%}    & 12.41\% \\
HyperFed             & 11.11\%    & 11.08\%    & 11.43\%    & 11.34\%    & 11.10\%    & 11.30\%    & 11.35\%    & \multicolumn{1}{c|}{11.32\%}    & 11.25\% \\ \hline
FedPalm              & -          & -          & -          & -          & -          & -          & -          & \multicolumn{1}{c|}{-}          & \textbf{4.27\%}  \\ \hline
\end{tabular}
}
\noindent\hspace*
{\dimexpr\oddsidemargin+1in\relax}%
\parbox{\textwidth}{FedPalm utilizes the open-set model exclusively in the open-set setting.}
\label{tab:op_pr_tongji}
\end{table*}

\begin{table}[!t]
\centering
\caption{The Open-Set EER Performance in Tongji Dataset (Generic FL Methods).}
\begin{tabular}{p{2.5cm}|>{\centering\arraybackslash}p{3cm}}
\hline
Method &  EER \\ \hline
\multicolumn{2}{c}{Non-FL Method}                       \\ \hline
w/o FL         & 10.89\%                       \\ \hline
\multicolumn{2}{c}{Generic FL Methods}                  \\ \hline
FedAvg         & 4.77\%                        \\
FedProx        & 5.18\%                        \\
FedNova        & 7.81\%                        \\
MOON           & 4.66\%                        \\
FedKD          & 5.21\%                        \\ \hline
FedPalm & \textbf{4.27}\% \\ \hline
\end{tabular}
\label{tab:op_ge_tongji}
\end{table}

\renewcommand{\arraystretch}{1.1}

\section{Experiment}
\subsection{Experimental Settings}
Our proposed method is implemented using the PyTorch framework and optimized with Adam \cite{kingma2014adam} at a learning rate of 0.01. The experimental environment in this study is as follows: AMD Ryzen 7 5800X CPU @3.80 GHz, 32 GB of RAM, and an NVIDIA GTX 3080 Ti.

We validated the proposed FedPalm on three public palmprint recognition datasets, including IITD~\cite{kumar2010personal}, Tongji~\cite{zhang2017towards}, and PolyU~\cite{zhang2003online}. For each dataset, we split the users into two halves: one half was used as the test set, while the remaining half served as the training set. Both open-set and closed-set scenarios were simulated. In the closed-set scenario, half of the data was used as the closed-set query set for the users assigned to the training set, while the remaining data was treated as closed-set gallery images. Therefore, the IDs do not overlap between the closed- and open-set settings.

We compared our FedPalm approach with both generic and personalized FL methods. For generic FL methods, we include FedAvg~\cite{mcmahan2017communication}, FedProx~\cite{li2020federated}, FedNova~\cite{wang2020tackling}, MOON~\cite{li2021model}, and FedKD~\cite{wu2022communication}. For personalized FL methods, we evaluated FedPer~\cite{arivazhagan2019federated}, FedBN~\cite{li2021fedbn}, and HyperFed~\cite{yang2023hypernetwork}. To ensure fairness, we kept the training settings consistent across all methods, with CompNet~\cite{liang2021compnet} used as the backbone.

In our experiments, we simulated an environment comprising eight clients, each assigned a distinct set of user IDs. To maintain this separation, the IDs in the training dataset were evenly distributed among the eight clients, preserving a clear distinction of identities across clients. This design aligns with the benchmark settings, providing a controlled and consistent evaluation framework.

\renewcommand{\arraystretch}{1.1}
\begin{table*}[!t]
\centering
\caption{The Closed-Set EER Performance in IITD.}
\resizebox{.85\textwidth}{!}{
\begin{tabular}{lccccccccc}
\hline
\multicolumn{1}{c}{} & Client \#1 & Client \#2 & Client \#3 & Client \#4 & Client \#5 & Client \#6 & Client \#7 & \multicolumn{1}{c|}{Client \#8} & Average \\ \hline
w/o FL               & 0.06\%     & 5.50\%     & 0.99\%     & 2.87\%     & 1.72\%     & 0.41\%     & 0.60\%     & \multicolumn{1}{c|}{1.79\%}     & 1.74\%          \\ \hline
\multicolumn{10}{c}{Generic FL Methods}                                                  \\
\hline 
FedAvg               & 0.23\%     & 4.13\%     & 3.45\%     & 5.75\%     & 2.83\%     & 0.57\%     & 0.88\%     & \multicolumn{1}{c|}{2.38\%}     & 2.53\%          \\
FedProx              & 1.16\%     & 5.17\%     & 4.02\%     & 6.32\%     & 4.47\%     & 1.15\%     & 1.46\%     & \multicolumn{1}{c|}{1.85\%}     & 3.20\%          \\
FedNova              & 17.82\%    & 15.89\%    & 18.39\%    & 19.62\%    & 17.82\%    & 20.11\%    & 19.64\%    & \multicolumn{1}{c|}{18.45\%}    & 18.47\%         \\
MOON                 & 0.39\%     & 4.02\%     & 3.45\%     & 5.75\%     & 2.30\%     & 0.66\%     & 0.79\%     & \multicolumn{1}{c|}{1.79\%}     & 2.39\%          \\
FedKD                & 0.51\%     & 4.02\%     & 3.45\%     & 5.75\%     & 2.30\%     & 0.57\%     & 0.60\%     & \multicolumn{1}{c|}{1.79\%}     & 2.37\%          \\
\hline
\multicolumn{10}{c}{Personazlied FL Methods} 
 \\   \hline
FedPer               &  0.14\%          &     2.30\%       &   2.30\%         &    5.17\%        &      1.72\%      &  1.07\%          &  1.09\%          & \multicolumn{1}{c|}{3.57\%}           &    2.18\%             \\
FedBN                & 1.77\%     & 6.90\%     & 7.47\%     & 6.69\%     & 2.81\%     & 5.75\%     & 5.40\%     & \multicolumn{1}{c|}{5.36\%}     & 5.27\%          \\
HyperFed             & 0.29\%     & 5.75\%     & 2.87\%     & 4.62\%     & 3.45\%     & 2.30\%     & 0.60\%     & \multicolumn{1}{c|}{2.98\%}     & 2.86\%          \\ \hline
FedPalm              & 0.02\%     & 2.30\%     & 0.57\%     & 3.45\%     & 2.16\%     & 1.09\%     & 0.09\%     & \multicolumn{1}{c|}{1.43\%}     & \textbf{1.39\%} \\ \hline
\end{tabular}
}
\label{tab:close_iitd}
\end{table*}

\begin{table*}[!t]
\centering
\caption{The Open-Set EER Performance in IITD Dataset (Personalized FL Methods).}
\resizebox{.85\textwidth}{!}{
\begin{tabular}{lccccccccc}
\hline
\multicolumn{1}{c}{Method} & Client \#1 & Client \#2 & Client \#3 & Client \#4 & Client \#5 & Client \#6 & Client \#7 & \multicolumn{1}{c|}{Client \#8} & Average  \\ \hline
\multicolumn{10}{c}{Non-FL Method}  \\ \hline
w/o FL               & 15.65\%    & 15.14\%    & 15.05\%    & 14.86\%    & 16.38\%    & 15.22\%    & 14.73\%    & \multicolumn{1}{c|}{13.77\%}    & 15.23\% \\ \hline
\multicolumn{10}{c}{Personalized FL Methods} \\ \hline
FedPer               & 14.78\%    & 14.64\%    & 13.04\%    & 14.06\%    & 14.64\%    & 15.16\%    & 13.85\%    & \multicolumn{1}{c|}{13.19\%}    & 14.17\% \\
FedBN                & 10.58\%    & 10.26\%    & 9.93\%     & 9.77\%     & 9.41\%     & 11.19\%    & 10.42\%    & \multicolumn{1}{c|}{9.57\%}     & 10.14\% \\
HyperFed             & 7.61\%     & 7.54\%     & 6.81\%     & 7.68\%     & 7.10\%     & 7.46\%     & 8.45\%     & \multicolumn{1}{c|}{7.90\%}     & 7.57\%  \\ \hline
FedPalm              & -          & -          & -          & -          & -          & -          & -          & \multicolumn{1}{c|}{-}          & \textbf{6.18\%} \\ \hline
\end{tabular}
}
\noindent\hspace*
{\dimexpr\oddsidemargin+1in\relax}%
\parbox{\textwidth}{FedPalm utilizes the open-set model exclusively in the open-set setting.}
\label{tab:op_pr_iitd}
\end{table*}

\begin{table}[!t]
\centering
\caption{The Open-Set EER Performance in IITD Dataset (Generic FL Methods).}
\begin{tabular}{p{2.5cm}|>{\centering\arraybackslash}p{3cm}}
\hline
Method &  EER \\ \hline
\multicolumn{2}{c}{Non-FL Method}                       \\ \hline
w/o FL         & 15.23\%                       \\ \hline
\multicolumn{2}{c}{Generic FL Methods}                  \\ \hline
FedAvg         & 7.06\%                        \\
FedProx        & 7.63\%                        \\
FedNova        & 13.45\%                       \\
MOON           & 7.25\%                        \\
FedKD          & 7.07\%                        \\ \hline
FedPalm & \textbf{6.18}\% \\ \hline
\end{tabular}
\label{tab:op_ge_iitd}
\end{table}

\subsection{Experiments}
For generic FL methods, we directly evaluate the shared model in both closed-set and open-set settings. However, personalized FL methods often overlook the challenge of handling unseen individuals. To address this, we average the performance of all local, personalized models to establish a baseline for the open-set setting. Their respective local models evaluate the closed-set setting for personalized FL models.


We then conduct a comprehensive evaluation of the closed-set and open-set performance of various methods across three public datasets, including Tongji (Tables~\ref{tab:cl_tongji},~\ref{tab:op_pr_tongji} and~\ref{tab:op_ge_tongji}), IITD~(Tables~\ref{tab:close_iitd},~\ref{tab:op_pr_iitd} and~\ref{tab:op_ge_iitd}), and PolyU (Tables~\ref{tab:close_polyu},~\ref{tab:op_pr_polyu} and~\ref{tab:op_ge_polyu}). The metric used is the equal error rate~(EER), a commonly used performance validation metric for palmprint recognition methods, representing the point on the ROC curve where the false acceptance rate~(FAR) equals the false rejection rate~(FRR), with $FRR = 1 - FAR$. Therefore, a lower EER indicates better performance. 

As shown in Table~\ref{tab:cl_tongji} to Table~\ref{tab:op_ge_polyu}, the proposed FedPalm consistently outperforms other FL methods across all datasets in both settings. Prior personalized FL methods typically excel in closed-set scenarios but suffer a significant performance drop when applied to unseen individuals in open-set settings. This limitation arises from their primary focus on local individuals, causing models to overfit specific client distributions and struggle with generalization. Consequently, these methods' performance gap between closed-set and open-set settings remains substantial. For closed-set scenarios, the w/o FL method can achieve performance comparable to personalized FL methods. However, its generalization ability is considerably weaker, making it insufficient to meet the demands of open-set scenarios. Therefore, the primary value of FL lies in enhancing robustness and generalization, which are crucial for biometric applications.

On the other hand, prior generic FL methods perform better in open-set settings but fail to deliver satisfactory results in closed-set settings. The root cause is data heterogeneity—these methods aim for an average-performing model across all clients, compromising their ability to adapt to local data distributions and leading to suboptimal outcomes for specific clients.

FedPalm addresses these limitations through a hybrid approach. It employs $N$ personalized closed-set models to adapt to local data distributions while leveraging a globally shared open-set model to enhance generalization. This design balances local specificity and global generalization, effectively mitigating challenges posed by non-IID data.  

For the closed-set models, FedPalm avoids integrating information from other clients, thus preserving the integrity of local distributions and ensuring convergence. The TEIM incorporates texture information from other texture experts to further enhance performance. TEIM selectively routes and integrates complementary texture features based on their similarity to the original features, which promotes feature diversity while minimizing conflicts arising from data heterogeneity.

The learnable parameters $\{\alpha, \beta\}$ of TEIM also enable adaptive weighting of original and side textural features, dynamically balancing local specialization and global generalization. This strengthens the model's discriminative power, improving its robustness to diverse data distributions and ensuring consistent performance in closed- and open-set settings.

By effectively addressing data heterogeneity and privacy constraints, FedPalm performs superiorly across various public datasets, setting a new standard in federated learning applications.

\renewcommand{\arraystretch}{1.1}
\begin{table*}[!t]
\centering
\caption{The Closed-Set EER Performance in PolyU.}
\resizebox{.85\textwidth}{!}{
\begin{tabular}{lccccccccc}
\hline
\multicolumn{1}{c}{} & Client \#1 & Client \#2 & Client \#3 & Client \#4 & Client \#5 & Client \#6 & Client \#7 & \multicolumn{1}{c|}{Client \#8} & Average \\ \hline
w/o FL               & 0.33\%     & 0.83\%     & 0.76\%     & 0.58\%     & 0.11\%     & 0.13\%     & 0.13\%     & \multicolumn{1}{c|}{0.09\%}     & 0.37\%  \\ \hline
\multicolumn{10}{c}{Generic FL Methods}                                                  \\ \hline 
FedAvg               & 1.13\%     & 1.25\%     & 0.79\%     & 1.11\%     & 0.33\%     & 0.17\%     & 0.40\%     & \multicolumn{1}{c|}{0.26\%}     & 0.67\%          \\
FedProx              & 1.33\%     & 2.25\%     & 1.83\%     & 2.00\%     & 1.33\%     & 0.26\%     & 0.13\%     & \multicolumn{1}{c|}{0.52\%}     & 1.20\%          \\
FedNova              & 4.74\%     & 5.54\%     & 6.02\%     & 5.83\%     & 5.25\%     & 4.30\%     & 5.00\%     & \multicolumn{1}{c|}{2.94\%}     & 4.95\%          \\
MOON                 & 1.20\%     & 1.29\%     & 0.79\%     & 1.04\%     & 0.35\%     & 0.20\%     & 0.27\%     & \multicolumn{1}{c|}{0.27\%}     & 0.68\%          \\
FedKD                & 1.17\%     & 1.25\%     & 0.75\%     & 1.08\%     & 0.39\%     & 0.18\%     & 0.26\%     & \multicolumn{1}{c|}{0.32\%}     & 0.67\%          \\
\hline
\multicolumn{10}{c}{Personazlied FL Methods} 
 \\   \hline
FedPer               & 0.36\%     & 0.42\%     & 0.13\%     & 1.13\%     & 0.12\%     & 0.09\%     & 0.52\%     & \multicolumn{1}{c|}{0.70\%}     & 0.43\%          \\
FedBN                & 0.67\%     & 0.75\%     & 1.33\%     & 1.71\%     & 0.79\%     & 0.87\%     & 1.00\%     & \multicolumn{1}{c|}{0.83\%}     & 0.99\%          \\
HyperFed             & 0.33\%     & 1.00\%     & 0.71\%     & 1.50\%     & 0.61\%     & 0.13\%     & 0.54\%     & \multicolumn{1}{c|}{1.17\%}     & 0.75\%          \\ \hline
FedPalm              & 0.42\%     & 0.17\%     & 0.42\%     & 0.17\%     & 0.08\%     & 0.43\%     & 0.01\%     & \multicolumn{1}{c|}{0.22\%}     & \textbf{0.24\%} \\ \hline

\end{tabular}
}
\label{tab:close_polyu}
\end{table*}

\begin{table*}[!t]
\centering
\caption{The Open-Set EER Performance in PolyU Dataset (Personalized FL Methods).}
\resizebox{.85\textwidth}{!}{
\begin{tabular}{lccccccccc}
\hline
\multicolumn{1}{c}{Method} & Client \#1 & Client \#2 & Client \#3 & Client \#4 & Client \#5 & Client \#6 & Client \#7 & \multicolumn{1}{c|}{Client \#8} & Average  \\ \hline
\multicolumn{10}{c}{Non-FL Method}  \\ \hline
w/o FL               & 12.55\%    & 12.49\%    & 12.71\%    & 12.48\%    & 13.38\%    & 13.10\%    & 12.56\%    & \multicolumn{1}{c|}{12.72\%}    & 12.75\% \\ \hline
\multicolumn{10}{c}{Personalized FL Methods} \\ \hline
FedPer               & 12.25\%    & 12.49\%    & 12.85\%    & 12.40\%    & 12.88\%    & 12.87\%    & 12.12\%    & \multicolumn{1}{c|}{13.53\%}    & 12.67\% \\
FedBN                & 7.04\%     & 7.03\%     & 6.92\%     & 7.16\%     & 6.81\%     & 7.08\%     & 6.93\%     & \multicolumn{1}{c|}{6.72\%}     & 6.96\%  \\
HyperFed             & 5.94\%     & 6.17\%     & 5.95\%     & 5.12\%     & 5.40\%     & 6.11\%     & 5.61\%     & \multicolumn{1}{c|}{5.91\%}     & 5.93\%  \\ \hline
FedPalm              & -          & -          & -          & -          & -          & -          & -          & \multicolumn{1}{c|}{-}          & \textbf{5.21\%} \\ \hline
\end{tabular}
}
\noindent\hspace*
{\dimexpr\oddsidemargin+1in\relax}%
\parbox{\textwidth}{FedPalm utilizes the open-set model exclusively in the open-set setting.}
\label{tab:op_pr_polyu}
\end{table*}

\begin{table}[!t]
\centering
\caption{The Open-Set EER Performance in PolyU Dataset (Generic FL Methods).}
\begin{tabular}{p{2.5cm}|>{\centering\arraybackslash}p{3cm}}
\hline
Method &  EER \\ \hline
\multicolumn{2}{c}{Non-FL Method}                       \\ \hline
w/o FL         & 12.75\%                       \\ \hline
\multicolumn{2}{c}{Generic FL Methods}                  \\ \hline
FedAvg         & 5.83\%                        \\
FedProx        & 5.74\%                        \\
FedNova        & 7.91\%                        \\
MOON           & 5.80\%                        \\
FedKD          & 5.77\%                        \\ \hline
FedPalm        & \textbf{5.21\%}               \\ \hline
\end{tabular}
\label{tab:op_ge_polyu}
\end{table}

\begin{figure*}
    \centering
    \includegraphics[width=.95\textwidth]{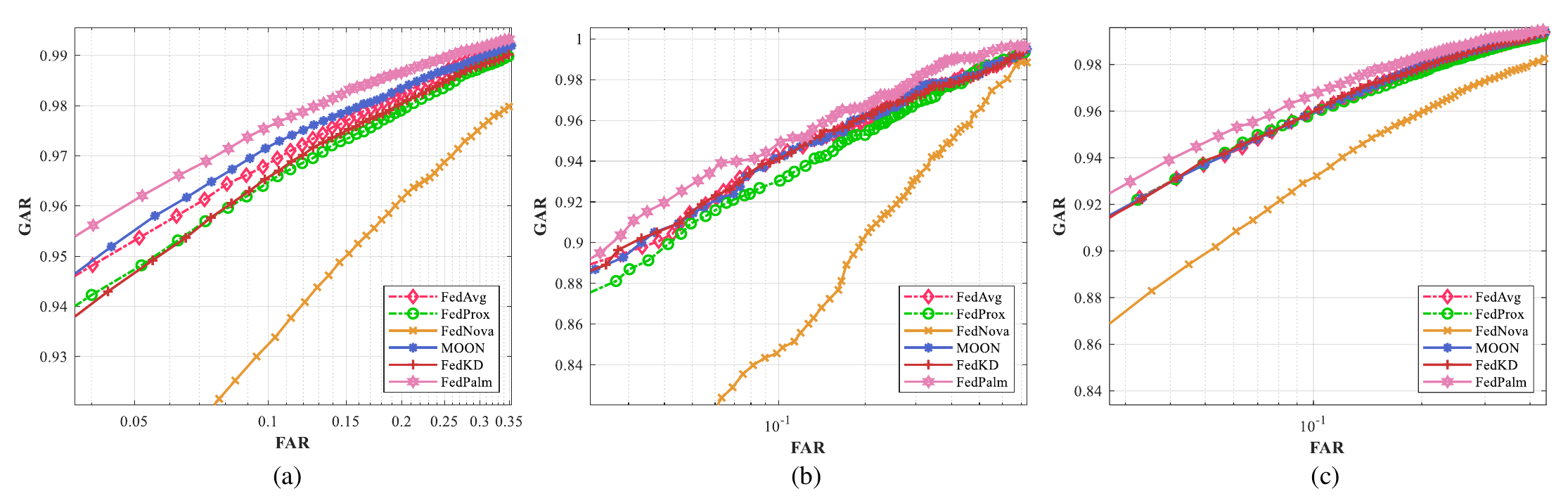}
    \caption{The ROC curves of different methods. (a)-(c) denote the Tongji, IITD, and PolyU results, respectively.}
    \label{fig:roc}
\end{figure*}

The open-set performance of various methods is assessed using the Receiver Operating Characteristic (ROC) curve, which illustrates the relationship between the Genuine Acceptance Rate (GAR) and the False Acceptance Rate (FAR). The ROC curve offers a detailed evaluation of trade-offs between these two metrics. We compared generic FL methods to ensure fairness, as personalized methods are not specifically designed for open-set scenarios. The results, depicted in Fig.~\ref{fig:roc}, reveal that the proposed FedPalm framework consistently delivers the best performance across all publicly available datasets.  

In addition to verification, we evaluated the identification capability of the proposed method, using Accuracy (ACC) as the metric. The results are presented in Table~\ref{tab:idn}. Given the difficulty plotting ROC curves for personalized FL methods, we employed the Area Under the Curve (AUC) as a quantifiable metric to assess the proposed method’s performance. The AUC results are also included in Table~\ref{tab:idn}. The data shows that FedPalm performs well in most cases, demonstrating its effectiveness.

\subsection{Ablation Study}

This subsection presents ablation studies to validate the effectiveness of the proposed components, with the results summarized in Table~\ref{tab:abl}. FedAvg serves as the baseline for our experiments. Specifically, "FedPalm*" represents the configuration where TEIM is incorporated from the beginning of training. In contrast, "FedPalm-1," "FedPalm-3," and "FedPalm-8" indicate settings where the parameter $K$ in Eq.~\eqref{eq:side} is set to 1, 3, and 8, respectively. These variations allow us to examine the impact of different configurations on performance.

\begin{table*}[!t]
\centering
\caption{The Identification Performance of Different Methods in IITD.}
\begin{tabular}{cc|ccccccccc|c}
\hline
                        &     & wo FL   & FedAvg  & FedProx & FedNova & MOON             & FedKD   & FedPer  & FedBN            & HyperFed & FedPalm          \\ \hline
\multirow{2}{*}{IITD}   & ACC & 70.16\% & 94.57\% & 94.13\% & 84.13\% & \textbf{94.78\%} & 94.13\% & 71.96\% & 93.26\%          & 93.04\%  & \textbf{94.78\%} \\
                        & AUC & 0.9246  & 0.9751  & 0.9772  & 0.9421  & 0.9753           & 0.9749  & 0.9324  & 0.9580           & 0.9742   & \textbf{0.9809}  \\ \hline
\multirow{2}{*}{Tongji} & ACC & 78.28\% & 97.07\% & 96.93\% & 92.17\% & 97.43\%          & 97.00\% & 79.93\% & 77.71\%          & 76.65\%  & \textbf{97.90\%} \\
                        & AUC & 0.9608  & 0.9888  & 0.9870  & 0.9746  & 0.9889           & 0.9871  & 0.9609  & 0.9486           & 0.9581   & \textbf{0.9906}  \\ \hline
\multirow{2}{*}{PolyU}  & ACC & 69.78\% & 96.19\% & \textbf{96.50\%} & 93.70\% & 96.09\%          & 96.14\% & 69.80\% & 96.49\% & 96.21\%  & 96.24\%          \\
                        & AUC & 0.9471  & 0.9852  & 0.9845  & 0.9728  & 0.9853           & 0.9855  & 0.9474  & 0.9800           & 0.9844   & \textbf{0.9879} \\ \hline
\end{tabular}
\label{tab:idn}
\end{table*}

\begin{table}[!t]
\centering
\caption{The Ablation Study of the Impact of $K$ in the TEIM on the IITD performance.}
\begin{tabular}{cccccc}
\hline
    & Baseline & FedPalm* & FedPalm-1 & FedPalm-3 & FedPalm-8 \\
    \hline
EER & 7.06\%   & 7.90\%   & 6.71\%    & \textbf{6.18\%}    & 7.89\%    \\
ACC & 94.57\%  & 90.87\%  & 93.91\%   & \textbf{94.78\%}   & 91.52\%   \\
AUC & 0.9751   & 0.9741   & 0.9793    & \textbf{0.9809}    & 0.9744   \\
\hline
\end{tabular}
\vspace{5pt}
\label{tab:abl}
\end{table}

As shown in Table~\ref{tab:abl}, introducing TEIM at the beginning of training yields suboptimal results. This is primarily because the textural experts have not yet stabilized, and the fluctuating TEIM inputs hinder the convergence of the training process. In contrast, selectively routing additional textural features through TEIM significantly improves performance, demonstrating the advantages of leveraging diverse expert features. However, routing all textural features results in performance degradation, likely due to the inclusion of irrelevant or conflicting information that weakens the discriminative power of the generated features.

These findings emphasize the importance of carefully selecting a limited number of relevant textural features to balance diversity and consistency. Among the configurations, "FedPalm-3" delivers the best performance, indicating that routing a moderate number of features enriches the textural representation while maintaining training stability. We recommend setting the hyperparameter $K$ to 3 for optimal results based on this analysis.




\section{Conclusion}

This paper introduces FedPalm, a unified federated learning framework for 
closed-set and open-set palmprint verification. By addressing privacy concerns and data heterogeneity, FedPalm ensures privacy-preserving model training without compromising performance. Leveraging personalized and global models, FedPalm balances local adaptability with global generalization. The proposed Textural Expert Interaction Module~(TEIM) enhances feature discrimination and robustness by dynamically integrating textural features. Extensive experiments on publicly available datasets demonstrate the superior performance of FedPalm, thereby validating its effectiveness. This framework establishes a benchmark for FL-based biometric systems, paving the way for future advancements in secure and generalizable palmprint recognition.

\bibliographystyle{ieeetr}
\bibliography{ref}

\begin{thebibliography}{10}

\bibitem{jain2004introduction}
A.~K. Jain, A.~Ross, and S.~Prabhakar, ``An introduction to biometric recognition,'' {\em IEEE Transactions on circuits and systems for video technology}, vol.~14, no.~1, pp.~4--20, 2004.

\bibitem{fei2018feature}
L.~Fei, G.~Lu, W.~Jia, S.~Teng, and D.~Zhang, ``Feature extraction methods for palmprint recognition: A survey and evaluation,'' {\em IEEE Trans. Syst. Man Cybern. Syst.}, vol.~49, no.~2, pp.~346--363, 2018.

\bibitem{zhang2003online}
D.~Zhang, W.-K. Kong, J.~You, and M.~Wong, ``Online palmprint identification,'' {\em IEEE Trans. Pattern Anal. Mach. Intell.}, vol.~25, no.~9, pp.~1041--1050, 2003.

\bibitem{jia2008palmprint}
W.~Jia, D.-S. Huang, and D.~Zhang, ``Palmprint verification based on robust line orientation code,'' {\em Pattern Recognit.}, vol.~41, no.~5, pp.~1504--1513, 2008.

\bibitem{yang20212TCC}
Z.~Yang, L.~Leng, T.~Wu, M.~Li, and J.~Chu, ``Multi-order texture features for palmprint recognition,'' {\em Artif. Intell. Rev.}, vol.~56, no.~2, pp.~995--1011, 2023.

\bibitem{xu2018drcc}
Y.~Xu, L.~Fei, J.~Wen, and D.~Zhang, ``Discriminative and robust competitive code for palmprint recognition,'' {\em IEEE Trans. Syst. Man Cybern. Syst.}, vol.~48, no.~2, pp.~232--241, 2018.

\bibitem{dian2016contactless}
L.~Dian and S.~Dongmei, ``Contactless palmprint recognition based on convolutional neural network,'' in {\em IEEE International Conference on Signal Processing (ICSP)}, pp.~1363--1367, IEEE, 2016.

\bibitem{liang2021compnet}
X.~Liang, J.~Yang, G.~Lu, and D.~Zhang, ``Compnet: Competitive neural network for palmprint recognition using learnable gabor kernels,'' {\em IEEE Signal Process. Lett.}, vol.~28, pp.~1739--1743, 2021.

\bibitem{yang2023ccnet}
Z.~Yang, H.~Huangfu, L.~Leng, B.~Zhang, A.~B.~J. Teoh, and Y.~Zhang, ``Comprehensive competition mechanism in palmprint recognition,'' {\em IEEE Trans. Inf. Forensics Secur.}, vol.~18, pp.~5160--5170, 2023.

\bibitem{yan2024toward}
L.~Yan, F.~Wang, L.~Leng, and A.~B.~J. Teoh, ``Toward comprehensive and effective palmprint reconstruction attack,'' {\em Pattern Recognition}, p.~110655, 2024.

\bibitem{voigt2017eu}
P.~Voigt and A.~Von~dem Bussche, ``The eu general data protection regulation (gdpr),'' {\em A Practical Guide, 1st Ed., Cham: Springer International Publishing}, vol.~10, no.~3152676, pp.~10--5555, 2017.

\bibitem{shao2024palmfedmetric}
H.~Shao, C.~Liu, X.~Li, and D.~Zhong, ``Privacy preserving palmprint recognition via federated metric learning,'' {\em IEEE Transactions on Information Forensics and Security}, vol.~19, pp.~878--891, 2024.

\bibitem{yang2024physics}
Z.~Yang, A.~B.~J. Teoh, B.~Zhang, L.~Leng, and Y.~Zhang, ``Physics-driven spectrum-consistent federated learning for palmprint verification,'' {\em International Journal of Computer Vision}, p.~4253–4268, 2024.

\bibitem{mcmahan2017communication}
B.~McMahan, E.~Moore, D.~Ramage, {\em et~al.}, ``Communication-efficient learning of deep networks from decentralized data,'' in {\em Artificial Intelligence and Statistics}, pp.~1273--1282, PMLR, 2017.

\bibitem{sabah2024model}
F.~Sabah, Y.~Chen, Z.~Yang, M.~Azam, N.~Ahmad, and R.~Sarwar, ``Model optimization techniques in personalized federated learning: A survey,'' {\em Expert Systems with Applications}, vol.~243, p.~122874, 2024.

\bibitem{fei2020feature}
L.~Fei, B.~Zhang, W.~Jia, J.~Wen, and D.~Zhang, ``Feature extraction for 3-d palmprint recognition: A survey,'' {\em IEEE Trans. Instrum. Meas.}, vol.~69, no.~3, pp.~645--656, 2020.

\bibitem{guo2009palmprint}
Z.~Guo, D.~Zhang, L.~Zhang, and W.~Zuo, ``Palmprint verification using binary orientation co-occurrence vector,'' {\em Pattern Recognit. Lett.}, vol.~30, no.~13, pp.~1219--1227, 2009.

\bibitem{kong2003palmprint}
W.~K. Kong, D.~Zhang, and W.~Li, ``Palmprint feature extraction using 2-d gabor filters,'' {\em Pattern Recognit.}, vol.~36, no.~10, pp.~2339--2347, 2003.

\bibitem{fei2016double}
L.~Fei, Y.~Xu, W.~Tang, and D.~Zhang, ``Double-orientation code and nonlinear matching scheme for palmprint recognition,'' {\em Pattern Recognit.}, vol.~49, pp.~89--101, 2016.

\bibitem{yang2020extreme}
Z.~Yang, L.~Leng, and W.~Min, ``Extreme downsampling and joint feature for coding-based palmprint recognition,'' {\em IEEE Transactions on Instrumentation and Measurement}, vol.~70, pp.~1--12, 2020.

\bibitem{fei2016half}
L.~Fei, Y.~Xu, and D.~Zhang, ``Half-orientation extraction of palmprint features,'' {\em Pattern Recognit. Lett.}, vol.~69, pp.~35--41, 2016.

\bibitem{zhong2018palmprint}
D.~Zhong, Y.~Yang, and X.~Du, ``Palmprint recognition using siamese network,'' in {\em Chinese Conference on Biometric Recognition~(CCBR)}, pp.~48--55, Springer, 2018.

\bibitem{chai2019boosting}
T.~Chai, S.~Prasad, and S.~Wang, ``Boosting palmprint identification with gender information using deepnet,'' {\em Future Gener. Comput. Syst.}, vol.~99, pp.~41--53, 2019.

\bibitem{genovese2019palmnet}
A.~Genovese, V.~Piuri, K.~N. Plataniotis, and F.~Scotti, ``Palmnet: Gabor-pca convolutional networks for touchless palmprint recognition,'' {\em IEEE Trans. Inf. Forensics Secur.}, vol.~14, no.~12, pp.~3160--3174, 2019.

\bibitem{jia2021deep}
W.~Jia, S.~Huang, B.~Wang, L.~Fei, Y.~Zhao, and H.~Min, ``Deep multi-loss hashing network for palmprint retrieval and recognition,'' in {\em Proc. IEEE Int. Joint Conf. Biom. (IJCB)}, pp.~1--8, IEEE, 2021.

\bibitem{wu2022multi}
T.~Wu, L.~Leng, and M.~K. Khan, ``A multi-spectral palmprint fuzzy commitment based on deep hashing code with discriminative bit selection,'' {\em Artif. Intell. Rev.}, pp.~1--18, 2022.

\bibitem{dong2022co}
X.~Dong, M.~K. Khan, L.~Leng, and A.~B.~J. Teoh, ``Co-learning to hash palm biometrics for flexible iot deployment,'' {\em IEEE Internet of Things Journal}, vol.~9, no.~23, pp.~23786--23794, 2022.

\bibitem{yang2023co3net}
Z.~Yang, W.~Xia, Y.~Qiao, Z.~Lu, B.~Zhang, L.~Leng, and Y.~Zhang, ``Co3net: Coordinate-aware contrastive competitive neural network for palmprint recognition,'' {\em IEEE Trans. Instrum. Meas.}, vol.~72, p.~3276506, 2023.

\bibitem{gao2024cross}
C.~Gao, Z.~Yang, T.-S. Ng, M.~Zhu, and A.~B.~J. Teoh, ``Cross-chirality palmprint verification: Left is right for the right palmprint,'' {\em arXiv preprint arXiv:2409.13056}, 2024.

\bibitem{yu2023multi}
L.~Yu, Q.~Yi, and K.~Zhou, ``Multi-scale discriminant representation for generic palmprint recognition,'' {\em Neural Computing and Applications}, vol.~35, no.~18, pp.~13147--13165, 2023.

\bibitem{li2020federated}
T.~Li, A.~K. Sahu, M.~Zaheer, {\em et~al.}, ``Federated optimization in heterogeneous networks,'' in {\em Pro. .Machine Learn. Syst.}, vol.~2, pp.~429--450, 2020.

\bibitem{wang2020tackling}
J.~Wang, Q.~Liu, H.~Liang, G.~Joshi, and H.~V. Poor, ``Tackling the objective inconsistency problem in heterogeneous federated optimization,'' in {\em Advances in neural information processing systems}, vol.~33, pp.~7611--7623, 2020.

\bibitem{li2021model}
Q.~Li, B.~He, and D.~Song, ``Model-contrastive federated learning,'' in {\em Proceedings of the IEEE/CVF Conference on Computer Vision and Pattern Recognitition}, pp.~10713--10722, 2021.

\bibitem{wu2022communication}
C.~Wu, F.~Wu, L.~Lyu, Y.~Huang, and X.~Xie, ``Communication-efficient federated learning via knowledge distillation,'' {\em Nat. Commun.}, vol.~13, no.~1, p.~2032, 2022.

\bibitem{zhang2023fedala}
J.~Zhang, Y.~Hua, H.~Wang, T.~Song, Z.~Xue, R.~Ma, and H.~Guan, ``Fedala: Adaptive local aggregation for personalized federated learning,'' in {\em Proceedings of the AAAI Conference on Artificial Intelligence}, vol.~37, pp.~11237--11244, 2023.

\bibitem{yang2023dynamic}
X.~Yang, W.~Huang, and M.~Ye, ``Dynamic personalized federated learning with adaptive differential privacy,'' in {\em Advances in Neural Information Processing Systems}, vol.~36, pp.~72181--72192, 2023.

\bibitem{arivazhagan2019federated}
M.~G. Arivazhagan, V.~Aggarwal, A.~K. Singh, and S.~Choudhary, ``Federated learning with personalization layers,'' {\em arXiv preprint arXiv:1912.00818}, 2019.

\bibitem{li2021fedbn}
X.~Li, M.~Jiang, X.~Zhang, {\em et~al.}, ``Fedbn: Federated learning on non-iid features via local batch normalization,'' {\em arXiv preprint arXiv:2102.07623}, 2021.

\bibitem{huang2023rethinking}
W.~Huang, M.~Ye, Z.~Shi, H.~Li, and B.~Du, ``Rethinking federated learning with domain shift: A prototype view,'' in {\em 2023 IEEE/CVF Conference on Computer Vision and Pattern Recognition (CVPR)}, pp.~16312--16322, IEEE, 2023.

\bibitem{cheng2023protohar}
D.~Cheng, L.~Zhang, C.~Bu, X.~Wang, H.~Wu, and A.~Song, ``Protohar: Prototype guided personalized federated learning for human activity recognition,'' {\em IEEE Journal of Biomedical and Health Informatics}, vol.~27, no.~8, pp.~3900--3911, 2023.

\bibitem{dai2023tackling}
Y.~Dai, Z.~Chen, J.~Li, S.~Heinecke, L.~Sun, and R.~Xu, ``Tackling data heterogeneity in federated learning with class prototypes,'' in {\em Proc. AAAI Conf. Artif. Intell. (AAAI)}, vol.~37, pp.~7314--7322, 2023.

\bibitem{yang2023hypernetwork}
Z.~Yang, W.~Xia, Z.~Lu, Y.~Chen, X.~Li, and Y.~Zhang, ``Hypernetwork-based physics-driven personalized federated learning for ct imaging,'' {\em IEEE Trans. Neural Netw. Learn. Syst.}, 2023.

\bibitem{chen2024hyperfednet}
X.~Chen, Y.~Huang, Z.~Xie, and J.~Pang, ``Hyperfednet: Communication-efficient personalized federated learning via hypernetwork,'' {\em arXiv preprint arXiv:2402.18445}, 2024.

\bibitem{shin2024effective}
Y.~Shin, K.~Lee, S.~Lee, Y.~R. Choi, H.-S. Kim, and J.~Ko, ``Effective heterogeneous federated learning via efficient hypernetwork-based weight generation,'' in {\em Proceedings of the ACM Conference on Embedded Networked Sensor Systems}, pp.~112--125, 2024.

\bibitem{shao2023privacy}
H.~Shao, C.~Liu, X.~Li, and D.~Zhong, ``Privacy preserving palmprint recognition via federated metric learning,'' {\em IEEE Transactions on Information Forensics and Security}, 2023.

\bibitem{shao2020towards}
H.~Shao and D.~Zhong, ``Towards privacy palmprint recognition via federated hash learning,'' {\em Electron. Lett.}, vol.~56, no.~25, pp.~1418--1420, 2020.

\bibitem{kingma2014adam}
D.~P. Kingma and J.~Ba, ``Adam: A method for stochastic optimization,'' {\em arXiv preprint arXiv:1412.6980}, 2014.

\bibitem{kumar2010personal}
A.~Kumar and S.~Shekhar, ``Personal identification using multibiometrics rank-level fusion,'' {\em IEEE Trans. Syst. Man. Cybern. c, Appl. Rev.}, vol.~41, no.~5, pp.~743--752, 2010.

\bibitem{zhang2017towards}
L.~Zhang, L.~Li, A.~Yang, Y.~Shen, and M.~Yang, ``Towards contactless palmprint recognition: A novel device, a new benchmark, and a collaborative representation based identification approach,'' {\em Pattern Recognit.}, vol.~69, pp.~199--212, 2017.

\end{thebibliography}



\begin{IEEEbiography}[{\includegraphics[width=1in,height=1.25in,clip,keepaspectratio]{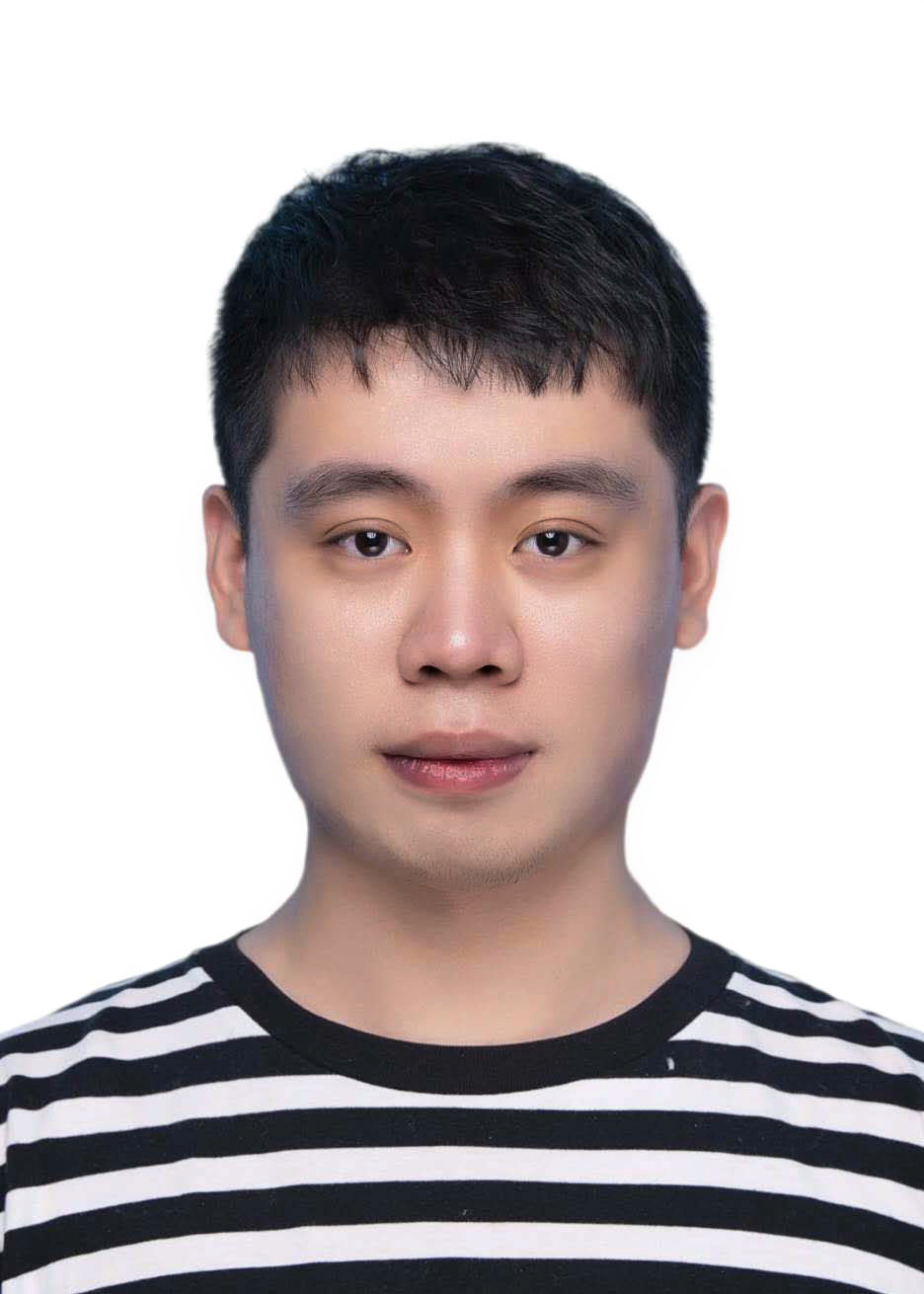}}]{Ziyuan Yang}
received an M.S. degree in computer science from the  School of Information Engineering, Nanchang University, Nanchang, China, in 2021. He is currently pursuing his Ph.D. degree with the College of Computer Science, Sichuan University, China. He is a visiting Ph.D. student at Centre for Frontier AI Research, Agency for Science, Technology and Research (A*STAR), Singapore. His research interests include biometrics, distributed learning, and security analysis.
\end{IEEEbiography}

\begin{IEEEbiography}[{\includegraphics[width=1in,height=1.25in,clip,keepaspectratio]{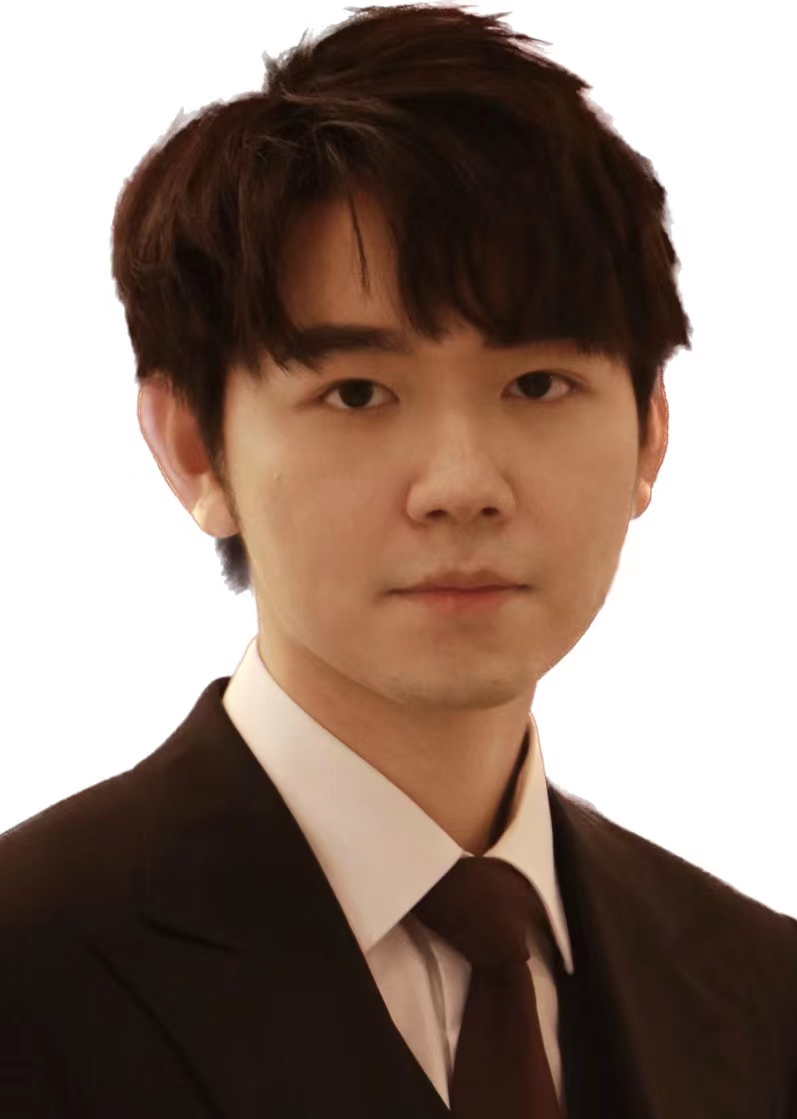}}]{Yingyu Chen}
received the B.S. degree in computer science from the School of Software, Sichuan University, Chengdu, China. He is currently pursuing his Ph.D. degree with the College of Computer Science, Sichuan University, China. His research interests include uncertainty quantification, machine learning and medical image analysis.
\end{IEEEbiography}

\begin{IEEEbiography}[{\includegraphics[width=1in,height=1.25in,clip,keepaspectratio]{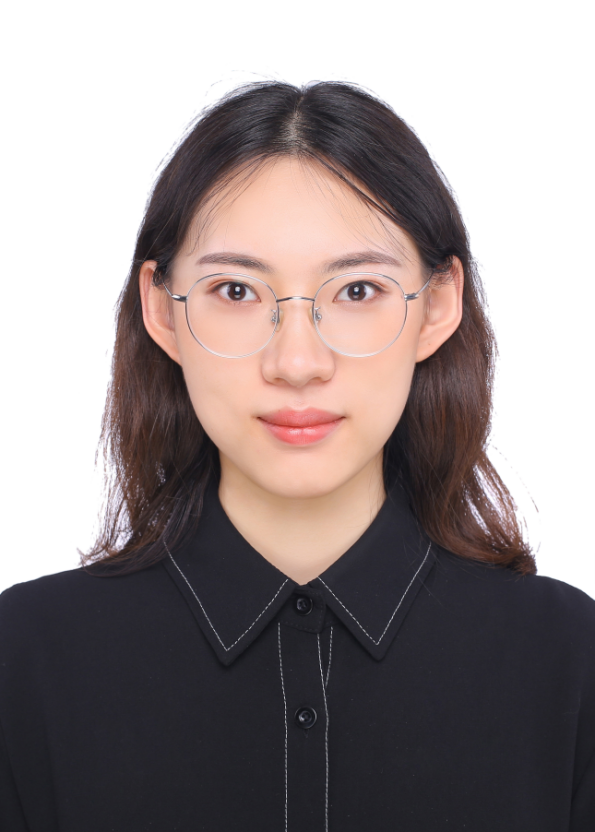}}]{Chengrui Gao}
received the M.S. degree from the School of Electronic Information, Sichuan University, Chengdu, China, in 2021. She is currently pursuing her Ph.D. degree at the College of Computer Science, Sichuan University. She was a joint Ph.D. student with Yonsei University supported by the China Scholarship Council Program in 2023. Her interests include biometrics and pattern recognition.
\end{IEEEbiography}

\begin{IEEEbiography}[{\includegraphics[width=1in,height=1.25in,clip,keepaspectratio]{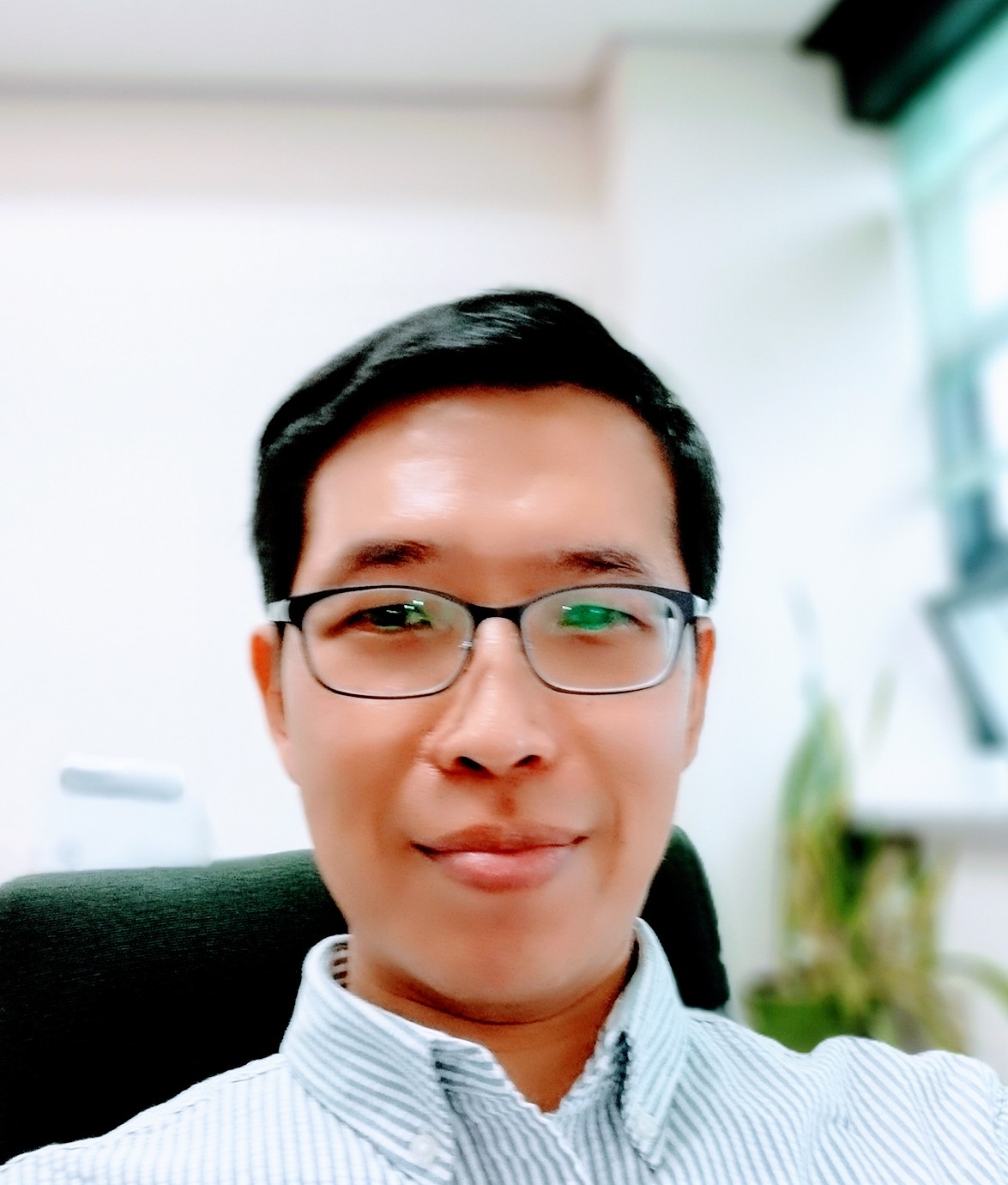}}]{Andrew Beng Jin Teoh}
(Senior Member, IEEE) received the B.Eng. degree in electronic and the Ph.D. degree from the National University of Malaysia in 1999 and 2003, respectively. He is currently a Full Professor with the Electrical and Electronic Engineering Department, College of Engineering, Yonsei University, South Korea. He has published over 350 international refereed journal articles and conference papers and edited several book chapters and volumes. His research, for which he has received funding, focuses on biometric applications and biometric security. His current research interests include machine learning and information security. He was the Guest Editor of the IEEE Signal Processing Magazine and a Senior Associate Editor of IEEE Transactions on Information Forensics and Security, IEEE Biometrics Compendium, and Machine Learning with Applications.
\end{IEEEbiography}

\begin{IEEEbiography}[{\includegraphics[width=1in,height=1.25in,clip,keepaspectratio]{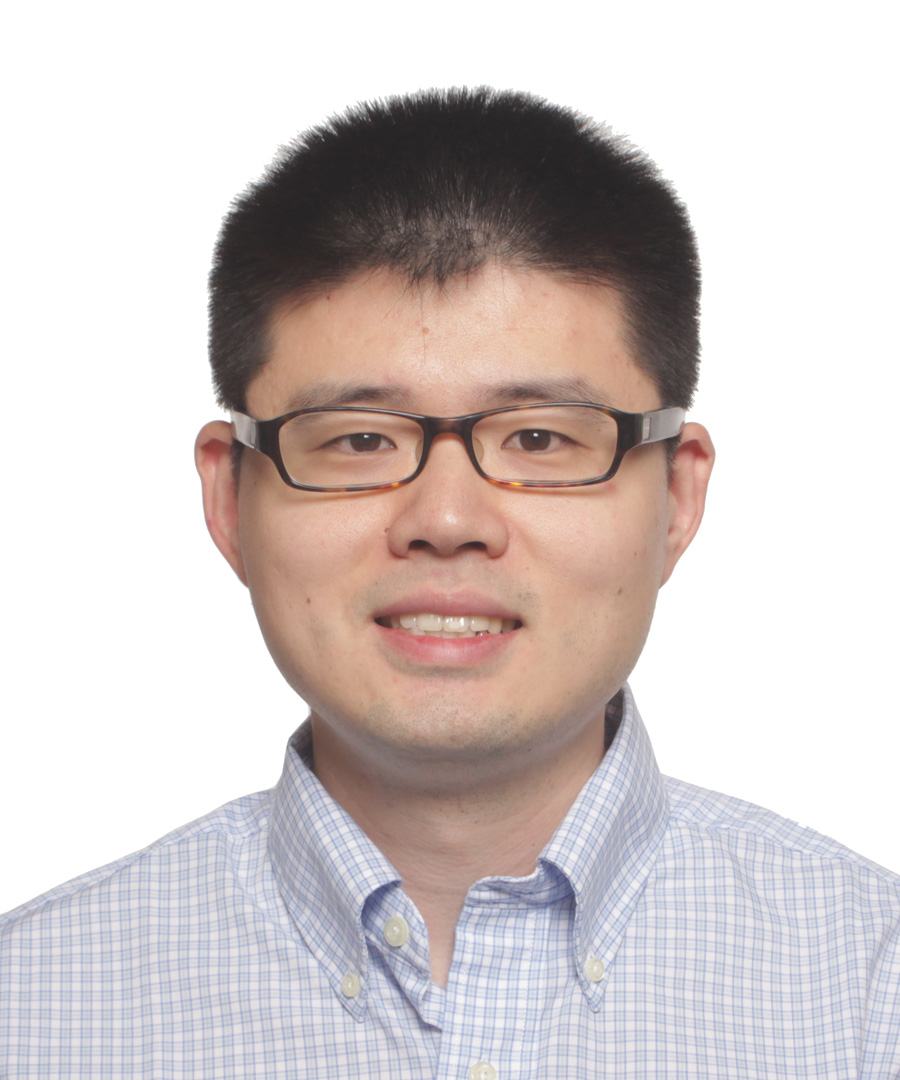}}]{Bob Zhang}
(Senior Member, IEEE) received the B.A. degree in computer science from York University, Toronto, ON, Canada, in 2006, the M.A.Sc. degree in information systems security from Concordia University, Montreal, QC, Canada, in 2007, and the Ph.D. in electrical and computer engineering from the University of Waterloo, Waterloo, ON, Canada, in 2011. After graduating from the University of Waterloo, he remained with the Center for Pattern Recognition and Machine Intelligence, and later, he was a Postdoctoral Researcher with the Department of Electrical and Computer Engineering, Carnegie Mellon University, Pittsburgh, PA, USA. He is currently an Associate Professor with the Department of Computer and Information Science, University of Macau, Macau. His research interests focus on biometrics, pattern recognition, and image processing. Dr. Zhang is a Technical Committee Member of the IEEE Systems, Man, and Cybernetics Society and an Associate Editor of IEEE TRANSACTIONS ON NEURAL NETWORKS AND LEARNING SYSTEMS, Artificial Intelligence Review, and IET Computer Vision.\end{IEEEbiography}

\begin{IEEEbiography}[{\includegraphics[width=1in,height=1.25in,clip,keepaspectratio]{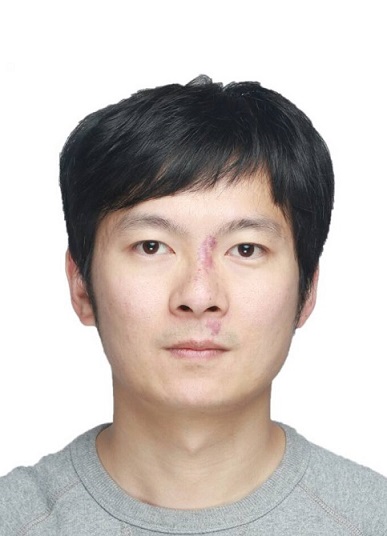}}]{Yi Zhang}
(Senior Member, IEEE) received the B.S., M.S., and Ph.D. degrees in computer science and technology from the College of Computer Science, Sichuan University, Chengdu, China, in 2005, 2008, and 2012, respectively. From 2014 to 2015, he was with the Department of Biomedical Engineering, Rensselaer Polytechnic Institute, Troy, NY, USA, as a Postdoctoral Researcher. He is currently a Full Professor with the School of Cyber Science and Engineering, Sichuan University, and is the Director of the deep imaging group (DIG). His research interests include medical imaging, compressive sensing, and deep learning. He authored more than 80 papers in the field of image processing. These papers were published in several leading journals, including IEEE TRANSACTIONS ON MEDICAL IMAGING, IEEE TRANSACTIONS ON COMPUTATIONAL IMAGING, Medical Image Analysis, European Radiology, Optics Express, etc., and reported by the Institute of Physics (IOP) and during the Lindau Nobel Laureate Meeting. He received major funding from the National Key R\&D Program of China, the National Natural Science Foundation of China, and the Science and Technology Support Project of Sichuan Province, China. He is a Guest Editor of the International Journal of Biomedical Imaging, Sensing and Imaging, and an Associate Editor of IEEE TRANSACTIONS ON MEDICAL IMAGING and IEEE TRANSACTIONS ON RADIATION AND PLASMA MEDICAL SCIENCES. 
\end{IEEEbiography}

\end{document}